\documentclass[letterpaper,10pt,conference]{ieeeconf}
\IEEEoverridecommandlockouts

\usepackage{iftex}
\ifPDFTeX
  \usepackage{graphicx}
\else
  \usepackage[dvipdfmx]{graphicx}
  \usepackage{bmpsize}
\fi
\usepackage{framed}
\usepackage{multirow}
\usepackage{booktabs}
\usepackage{url}
\usepackage{cite}
\usepackage{fancyvrb}
\usepackage{fvextra}
\usepackage{amsmath,amssymb,amsfonts}
\fvset{
  breaklines=true,
  breakanywhere=true,
  breaksymbolleft=\(\),
  breaksymbolsep=0pt
}
\usepackage[T1]{fontenc}
\usepackage{listings}
\begin{document}

\title{Breaking the 15\% Barrier: A Real-World Data-Driven System \\ for Proactive Social Robot Triggered by User Nonverbal Cues}

\author{Yuga Yano$^{1,2}$, Yuki Okafuji$^{2,3}$, Ryo Miyoshi$^{2,3}$, Sanae Yamashita$^{2,3}$, Yoshiki Ohira$^{2,3}$
   \thanks{$^{1}$ Kyushu Institute of Technology, Fukuoka, Japan}
   \thanks{$^{2}$ CyberAgent, Tokyo, Japan}
   \thanks{$^{3}$ The University of Osaka, Osaka, Japan}
   \thanks{For correspondence:~{\tt\small\ yano.yuuga158@mail.kyutech.jp}}
}
\date{November 2025}

\maketitle
\thispagestyle{empty}
\pagestyle{empty}

\begin{abstract}


Service robots in retail stores increasingly rely on cascaded speech pipelines (STT–LLM–TTS), yet many customer–robot interactions are initiated or guided by nonverbal behaviors such as approaching, waving, pointing, or showing items.
This paper studies such cues in a real-world store deployment with a teleoperated humanoid robot and shows that a non-negligible portion of robot turns are triggered by nonverbal behaviors rather than spoken input, revealing a limitation of audio-only dialogue systems.
In a 6-day in-the-wild deployment, 15.3\% of robot utterances were initiated by users' nonverbal behaviors rather than spoken input.
Based on an analysis of observed customer behaviors, we define a set of frequent, service-relevant nonverbal cues and develop a real-time multi-person, multi-label recognizer that runs online from video.
We then propose a dialogue framework that conditions LLM-based utterance generation on recognized nonverbal cue tokens, and optionally leverages a vision–language model when items are shown, enabling proactive robot responses without hand-crafted rules.
We evaluate the approach offline on nonverbal-triggered turns and demonstrate an online prototype that reacts to users' nonverbal cues in real time.
\end{abstract}


\section{Introduction}

Service robots capable of autonomous customer-service dialogue are being introduced in retail stores to reduce the workload of store staff and improve the customer experience~\cite{yano2024unified,inoue2025vap,okafuji2024eat}.
Recently, advances in large language models (LLMs) such as ChatGPT~\cite{hurst2024gpt4o} and Gemini~\cite{comanici2025gemini}, as well as speech-to-text (STT) and text-to-speech (TTS) technologies, have significantly improved the verbal interaction capabilities of service robots.
Meanwhile, many nonverbal behaviors---such as waving and pointing---often trigger interactions~\cite{urakami2023nonverbal}.
In contrast, widely deployed spoken dialogue systems are often implemented as a cascaded pipeline: the input speech is transcribed by STT, the transcript is fed into an LLM to generate a response, and the response is synthesized by TTS~\cite{chen2024dvoiceassit}.
In such systems based on only audio input, spoken dialogue systems cannot incorporate users' nonverbal behaviors as inputs, which leads to missed opportunities for engagement.

While many studies of human activity recognition have focused on developing dedicated models, recent advances in large-scale infrastructure models have made it possible to describe detailed user behavior using vision language models (VLMs)\cite{pmlr25ex-vad,xu2025streamingvlm,yuan2025videorefer}. However, such VLM-based approaches require a high-performance GPU for real-time recognition, and significant challenges remain before deployment as a customer-service dialogue system.
In this study, real-time recognition refers to a speed of at least 1 fps. This is because humans generally expect a response delay of approximately one second to user actions~\cite{Shiwa2008quick}; therefore, human activity recognition must be completed within this time frame, even when the time required for speech generation and voice synthesis is excluded.

\begin{figure}
    \centering
    \includegraphics[width=1.0\linewidth]{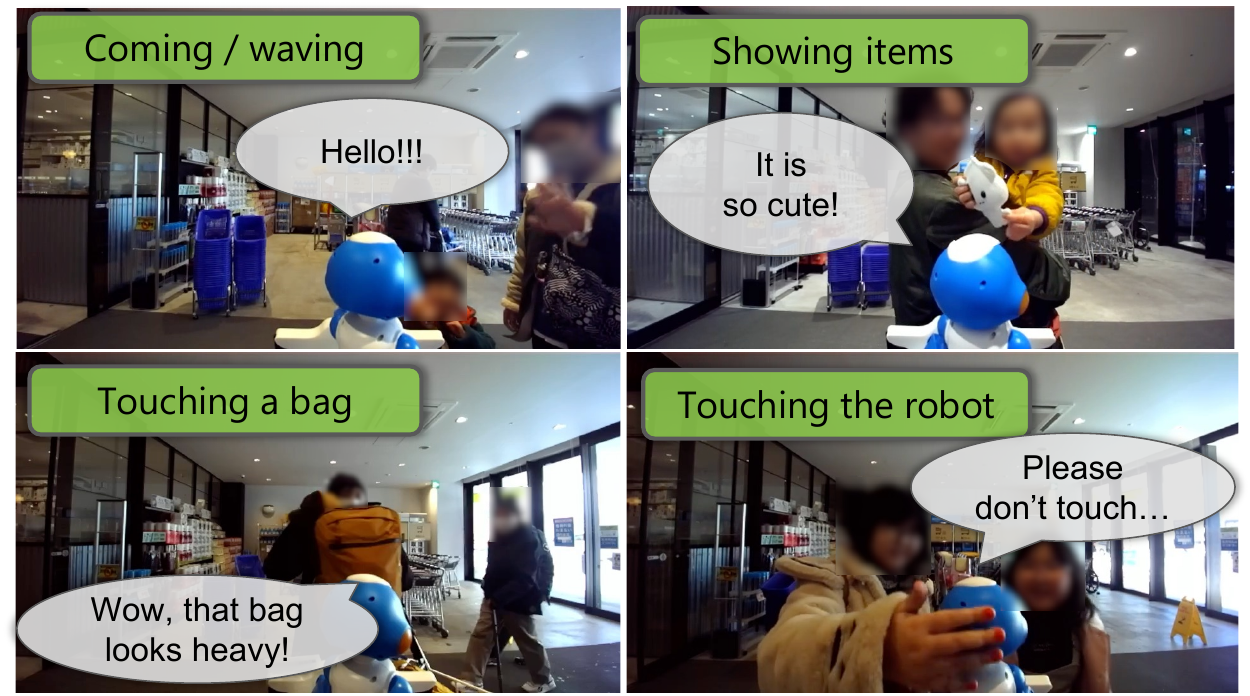}
    \caption{In this study, we (a) collect dialogue data from a remote-controlled robot installed in a physical store, (b) analyze utterances resulting from the user's nonverbal behavior, and (c) propose an autonomous dialogue system based on that analysis. We build a dialogue robot that can respond to the user's nonverbal behavior in real time.}
    \label{fig:teaser}
\end{figure}

Therefore, it is essential to analyze nonverbal behaviors that play important roles in customer-service interactions and to develop a dialogue framework that integrates a real-time recognition model.
Fig.~\ref{fig:teaser} provides an overview of our research objective.
In this paper, we aim to realize natural spoken interaction in real-world service environments.
To this end, we make two main contributions:

\begin{itemize}
  \item We collect human--robot interaction data in a physical retail store through teleoperation and analyze the robot's utterances. In particular, we quantify response patterns to users' spoken utterances and to users' nonverbal behaviors that trigger speech. We identify user nonverbal behaviors that influence dialogue and present an analysis that connects these findings to model design.
  \item We develop a real-time multi-label recognition model that detects important user nonverbal behaviors. We further propose an example of a dialogue framework that generates proactive responses based on the recognized behaviors. We demonstrate that the proposed system generates dialogue content conditioned on users' nonverbal behaviors. It matches human teleoperators' intent on frequent social cues, while rarer task-related directives remain a limitation.

\end{itemize}


In our first contribution, we deployed a teleoperated customer-service robot at the entrance of a retail store and categorized its utterances into responses to users' speech and utterances triggered by users' nonverbal behaviors. The latter cannot be handled by conventional speech-recognition-based dialogue systems combined with LLMs. Our analysis showed that about 15\% of the robot's utterances were nonverbal-triggered. From the 21 triggering behaviors identified, we selected nine as particularly important based on their frequency and relevance in service interactions.

In our second contribution, we develop a recognition model for the nine key behaviors and integrate it into a dialogue system. The model performs real-time multi-label prediction for multiple users. We also propose a dialogue framework that incorporates recognized behaviors in real time to generate proactive responses. Experimental results show that the system matches the teleoperator's intent on frequent social cues, while task-related directives remain a limitation.

\section{Related Works}

\subsection{Classification of User Nonverbal Behaviors in HRI}
In Human--Robot Interaction (HRI), users' attention, engagement, intention to participate, and turn-taking timing are often conveyed through nonverbal cues such as gaze, posture, interpersonal distance, gestures, and object manipulation. Because speech alone cannot capture these signals well, analyzing user nonverbal behavior has been a longstanding topic in HRI~\cite{urakami2023nonverbal}. In field settings (e.g., public spaces and retail stores), user motivations and engagement styles vary widely~\cite{amy2025hri_motivation}; therefore, it is important to clarify which directly observable behaviors in real environments should be treated as dialogue-relevant inputs.

In this study, we focus on observable behaviors rather than concepts that require estimating internal states such as emotions. Under this definition, nonverbal behaviors in HRI can be organized into the following observable units~\cite{urakami2023nonverbal}.

\begin{enumerate}
  \item \textbf{Proxemics / Approach--Avoid}:
    Changes in relative position and distance (e.g., approaching, stopping, staying nearby, or leaving) provide strong cues for interaction onset/offset and intervention timing. Distance interpretation depends on operational definitions for safety and comfort~\cite{neef2023appropriate}.
  \item \textbf{Body orientation / Posture}:
    Body, shoulder, and head orientation, as well as leaning forward, are interpretable indicators of attention and partner selection. This modality is often combined with other modalities to detect user intent.
  \item \textbf{Gaze / Head movement}:
    Gaze fixation, mutual gaze, and nodding are frequently used as backchannels and turn-transition cues. Recent work also adapts turn-taking models to HRI to improve fluency and inform when a robot should intervene~\cite{skantze2025turntaking}.
  \item \textbf{Gesture / Signaling}:
    Raising a hand, waving, and pointing are often linked to attention-seeking, engagement requests, and instructions. Relevant gesture sets vary by task and context and are usually defined for specific research aims.
  \item \textbf{Object manipulation / Showing}:
    Showing items or using a smartphone can reveal what the user is interested in. In customer service, attraction and engagement should be treated as a continuous process, making such behaviors important to capture~\cite{song2025robotclerk}.
  \item \textbf{Observable overt affect-like behavior}:
    Smiling or laughing can provide coarse cues of interaction success and acceptance when treated as observable behaviors rather than as direct estimates of internal states.
\end{enumerate}

While prior works provide rich classifications of nonverbal behavior, the findings are often tied to specific robot roles and environments. As a result, they are not yet systematically organized from a system-design viewpoint---especially regarding which observable behaviors should be prioritized for recognition when deploying dialogue systems in real-world settings. Choosing an appropriate granularity and coverage of behaviors as dialogue inputs requires prioritization along two axes: system feasibility (e.g., real-time recognition) and interaction value (e.g., intervention effectiveness).

\subsection{User Action Recognition}
Understanding customer behavior in retail environments relates to video-based human action understanding in computer vision, including action recognition and temporal action detection.
Many recent approaches leverage video features and vision--language models.

SlowFast~\cite{feichtenhofer2019slowfast} introduced a dual-pathway architecture that jointly models local motion and global context by combining a low-frame-rate stream for appearance/context with a lightweight high-frame-rate stream for motion.
Action Transformer~\cite{girdhar2019video} uses each person's ROI features as queries to aggregate spatiotemporal context from an entire clip for multi-label person action classification.
TubeR~\cite{zhao2022tuber} further introduces person-specific queries to predict trajectories and action labels across frames.
Beyond these performance-oriented methods, foundation models have enabled more fine-grained understanding.

ActionCLIP~\cite{wang2021actionclip} frames action recognition as video--text matching and exploits label semantics for zero-/few-shot transfer via prompting. It presented a pipeline that adapts pretrained image-/video--text models to action recognition through prompting.
VideoRefer~\cite{yuan2025videorefer} introduces an object-referring dataset/benchmark and a video LLM that models region and temporal representations for object-level spatiotemporal understanding.
By using SAM2~\cite{ravi2024sam2} for mask generation and tracking, user-specified regions can be input to support detailed description and reasoning.

Despite strong accuracy, foundation-model-based approaches often require substantial computation and inference time, making direct deployment in HRI difficult.
Store robots require low-latency estimation of real-time nonverbal behaviors for multiple people and dialogue generation based on those estimates.
We therefore propose a pipeline that leverages foundation-model representations while enabling fast multi-person, multi-label action classification and behavior-conditioned dialogue generation under HRI constraints.
Our target is real-time, multi-person recognition within the one-second response budget.
Large offline video models on server-class GPUs operate at a different point, so we prioritize deployability over offline accuracy rather than benchmarking against them (see Sec.~\ref{sec:action_recognition} for measured VLM latencies).

\subsection{Utterance Generation Based on Nonverbal Behavior}
Recent studies have adapted robot dialogue by recognizing user nonverbal behavior and using the results for utterance selection. Here, we focus on work where user nonverbal information affects utterance content (e.g., paraphrasing, continuation, or repair), not only turn-taking timing.

Constantin et al. observed that pointing often introduces referential ambiguity and proposed a system that estimates pointing and the pointed object to adapt utterances to the ambiguity level~\cite{constantin2023interactive}.
Axelsson and Skantze presented a robot that recognizes multimodal user responses (gaze, head movement, brief verbal feedback) and adjusts content using a knowledge graph updated by these recognition states~\cite{axelsson2023doyoufollow}.
More recently, LLMs have been integrated into dialogue systems: Allgeuer et al. incorporate nonverbal perception (e.g., posture or gesture detection) as textual states to generate responses consistent with user behavior~\cite{allgeuer2024chatty}.

These studies typically assume a small set of task-specific nonverbal behaviors. In customer service, however, social and task functions coexist, and it is unclear which observable behaviors drive which dialogue functions and to what extent.
Therefore, we analyze interaction data from a teleoperated dialogue robot deployed in a physical retail store.
We extract utterances triggered by user nonverbal behaviors and map them to a compact set of dialogue-function labels, then quantify the correspondence between behaviors and dialogue functions.
Based on this analysis, we build LLM-based utterance generation conditioned on the recognized behaviors.

\section{Conversation Data Collection and Analysis in A Real-world Store}
\label{sec:datacollect}

\subsection{Data Collection}
To analyze users' nonverbal behaviors in relation to spoken content in human--robot customer-service conversations, we collected data in a real retail store with approval from the ethical committee of The University of Osaka.

The dataset was recorded at a drugstore in Japan, where a teleoperated customer-service robot provided store information and small talk to customers. The robot was operational for a total of 32 hours (6 days), of which only 1 hour and 42 minutes were annotated, representing all interaction time where dialogue occurred.
%

\begin{figure}
    \centering
    \includegraphics[width=0.98\linewidth]{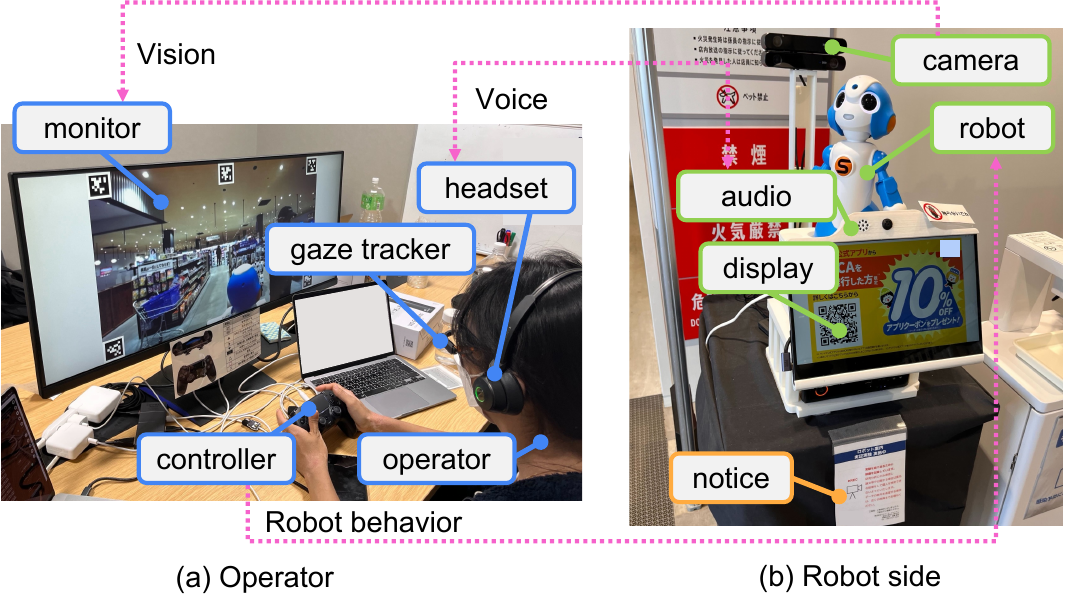}
    \caption{Data collection setup in the store. The operator viewed the rear-camera and controlled the robot's speech, gestures, and gaze behavior.}
    \label{fig:dataset_collection}
\end{figure}

Fig.~\ref{fig:dataset_collection} shows the teleoperation setup. We used Sota, a desktop-sized humanoid robot (approx.\ 0.3\,m tall). It has 8 degrees of freedom (arms, torso, neck). A rear camera streamed the scene to the operator in real time, and the robot used an onboard microphone and speaker for spoken dialogue. A small display in front of the robot presented guidance about the store's information.

The remote operator viewed the rear-camera feed and controlled the robot's speech using their own voice after conversion, gestures via a controller, and gaze via an eye-tracking device. To preserve the operator's customer-service skills, the robot had no autonomous functions; all behaviors were operator-controlled. The operator's voice was converted to a robot-like voice using MorphVOX to reduce an uncanny mismatch between appearance and voice~\cite{mitchel2011uncanny}. Using a DualSense wireless controller, the operator could trigger five gestures (raise a hand, wave goodbye, extend a hand, shake the head, and a shy motion), chosen with reference to prior work on data-driven co-speech gesture generation~\cite{nyatsanga2023gesture}. The operator's gaze was measured with Pupil Core and transmitted to the robot, which then oriented its gaze toward the user's face or body. Together, these components supported natural interaction through the robot.

\subsection{Annotation}
\label{ss:satsudora_annotation}

We annotated the collected dialogue data for the analyses and the experiments reported later. Four annotators with no conflict of interest performed the work. Label definitions and decision criteria were aligned with the researchers in advance, and all labels were finalized after double-checking by another annotator and/or a researcher.

We first transcribed the robot (operator)'s and customers' utterances as timestamped dialogue histories. By aligning transcripts with person-tracking IDs from the videos, we can attribute each utterance to a specific user.

We then classified all of the robot's utterances into responses to customers' spoken utterances and utterances initiated in reaction to customers' nonverbal behaviors (nonverbal-triggered utterances). A nonverbal-triggered utterance is a proactive robot utterance driven not by the customer's spoken content itself but by observable behaviors such as gaze, approach, or gestures. For nonverbal-triggered utterances, we further assigned two labels a) and b).

\paragraph{Categorization of user nonverbal behaviors that triggered the utterance}

This label captures the customer's behavior immediately before a nonverbal-triggered utterance. We designed labels by referring to the higher-level observational units 1)--6) in Related Work and selecting behaviors that are common in retail settings and likely to affect customer-service dialogue design and generation~\cite{urakami2023nonverbal}.

\paragraph{Categorization of the intent of nonverbal-triggered utterances}

This label classifies the function/intent of robot utterances into five categories. Based on prior frameworks for dialogue act schemas and task-oriented vs.\ casual conversation~\cite{bunt2020iso, chen2017survey}, we condensed teleoperated utterances in real-store customer-service scenes into a compact set that can be analyzed and mapped to implementation at the utterance level. Motivated by work on proactive dialogue strategies and interaction breakdown/repair~\cite{nothdurft2015proactive, tisserand2024unraveling}, we chose a granularity that accommodates co-occurring social obligations (e.g., greetings), task completion, engagement initiation, and user-state comments. To reduce interpretive variance, multiple authors finalized these labels by consensus.

\begin{itemize}
  \item \textbf{Social \& Greeting}: Greetings, farewells, thanks, and self-introductions (e.g., ``Hello,'' ``See you,'' ``I'm Sota.'')
  \item \textbf{Task-related info \& directives}: Store-related information, invitations, or directives (e.g., sales, app guidance, navigation, cautions; ``I recommend the app.'')
  \item \textbf{User/state comment \& empathy}: Comments or empathetic remarks about the user's behavior/appearance/belongings, weather, or condition (e.g., ``Your red coat looks nice,'' ``That bag looks heavy.'')
  \item \textbf{Engagement invitation}: Utterances that initiate or encourage conversation, including small-talk questions (e.g., ``Let's talk,'' ``Can I help you find something?'')
  \item \textbf{Other}: Utterances that do not fit the above categories
\end{itemize}

\subsection{Analysis Results}
We extracted all utterances of the operator and labeled each utterance as either a response to a customer utterance or a nonverbal-triggered utterance. Among 2,178 robot utterances, 334 were nonverbal-triggered utterances (15.3\%), indicating that a meaningful portion of utterance generation in customer-service interactions is driven by customers' nonverbal behaviors rather than speech input.

We then aggregated the customer behaviors that triggered these utterances and identified 21 behavior types. Table~\ref{tab:nonverbal-stats} summarizes their frequencies. The most frequent were \textit{Approached}, \textit{Watching from a distance}, and \textit{Waved}. These dominant behaviors align with prior findings that proxemics (distance/approach) and bodily cues of attention and interest (e.g., gaze and hand waving) are key signals for interaction initiation and engagement formation in public and service contexts~\cite{urakami2023nonverbal, neef2023appropriate, satake2009approach, chamoto2023task}. Note that the total count in Table I is larger than the number of nonverbal-triggered utterances because multiple nonverbal behaviors can co-occur (e.g., a customer approaches while waving) and be attributed to a single utterance.

From the 21 behaviors, we selected nine as \emph{important} for subsequent recognition: \textit{Approached}, \textit{Walked away}, \textit{User--user interaction}, \textit{Nodded}, \textit{Showed items}, \textit{Peered into robot}, \textit{Touched belongings}, \textit{Pointed}, and \textit{Waved}. These were chosen because they are relatively frequent, characteristic of customer-service scenes, and expected to be stably detectable in video. We intentionally retain \textit{Touched belongings} despite its low frequency because it is strongly associated with task-related guidance and represents a high-value intervention in retail service. We use these nine behaviors as the target recognition classes in the following sections.

We also mapped nonverbal-triggered utterances to the five intent categories (Table~\ref{tab:trigger-label-stats}). The most frequent intent was \textit{Social \& Greeting}, suggesting that operators often initiated greetings based on customers' nonverbal cues at interaction start. The second most frequent was \textit{Engagement invitation}, which tends to appear when customers remain silent or wait, implying that operators rely on nonverbal behaviors to decide when to ask follow-up questions or shift topics.

Finally, Table~\ref{tab:nonverbal-cross} summarizes how triggering behaviors relate to utterance intent. \textit{Approached}, \textit{Walked away}, \textit{Pointed}, and \textit{Waved} mainly triggered \textit{Social \& Greeting}, whereas \textit{Touched belongings} more often led to \textit{Task-related info \& directives}. \textit{User--user interaction} and \textit{Nodded} showed higher proportions of \textit{Engagement invitation}, and \textit{Showed items} tended to correspond to \textit{User/state comment \& empathy}. This correspondence helps characterize which intents are likely elicited by each behavior and informs design guidelines for utterance selection based on nonverbal behavior recognition.

\begin{table}[t]
  \centering
  \caption{Counts and percentages by nonverbal behavior category}
  \label{tab:nonverbal-stats}
  \begin{tabular}{lrr}
    \toprule
    Users' Nonverbal Behavior & Count & Percentage (\%) \\
    \midrule
    Approached & 118 & 29.2 \\
    Watching from a distance & 48 & 11.9 \\
    Waved & 47 & 11.6 \\
    Walked away & 32 & 7.9 \\
    Wandering around robot & 25 & 6.2 \\
    Touched / attempted to touch & 19 & 4.7 \\
    Walking while looking at robot & 17 & 4.2 \\
    Walking without looking at robot & 16 & 4.0 \\
    User--user interaction & 16 & 4.0 \\
    Peered into robot & 11 & 2.7 \\
    Rock-paper-scissors / hand gestures & 10 & 2.5 \\
    Pointed & 9 & 2.2 \\
    Other behaviors & 8 & 2.0 \\
    Using hand sanitizer & 6 & 1.5 \\
    Laughed & 6 & 1.5 \\
    Showed items & 5 & 1.2 \\
    Raised hand & 4 & 1.0 \\
    Nodded & 4 & 1.0 \\
    Touched belongings & 3 & 0.7 \\
    \midrule
    Total & 404 & 100.0 \\
    \bottomrule
  \end{tabular}
\end{table}

\begin{table}[t]
  \centering
  \caption{Counts and percentages by nonverbal-triggered utterance label}
  \label{tab:trigger-label-stats}
  \begin{tabular}{lrr}
    \toprule
    Nonverbal-triggered utterance label & Count & Percentage (\%) \\
    \midrule
    Social \& Greeting & 147 & 44.0 \\
    Engagement invitation & 100 & 29.9 \\
    User/state comment \& empathy & 45 & 13.5 \\
    Task-related info \& directives & 35 & 10.5 \\
    Other & 7 & 2.1 \\
    \midrule
    Total & 334 & 100.0 \\
    \bottomrule
  \end{tabular}
\end{table}

\begin{table*}[t]
  \centering
  \caption{Counts of utterance-intent labels for each of the nine target nonverbal behaviors (percentages are within each behavior).}
  \label{tab:nonverbal-cross}
  \begin{tabular}{l|cccc}
    \toprule
      & Social \& Greeting
      & Engagement invitation
      & User/state comment \& empathy
      & Task-related info \& directives \\
    \midrule
    Approached
      & 73 (66.4\%) & 26 (23.6\%) & 4 (3.6\%)  &  6 (5.5\%) \\
    Waved
      & 29 (65.9\%) &  9 (20.5\%) & 4 (9.1\%)  &  2 (4.6\%) \\
    Walked away
      & 22 (73.3\%) &  3 (10.0\%) & 2 (6.7\%)  &  1 (3.3\%) \\
    User--user interaction
      &  3 (23.1\%) &  5 (38.5\%) & 2 (15.4\%) &  3 (23.1\%) \\
    Peered into robot
      &  3 (27.3\%) &  4 (36.4\%) & 3 (27.3\%) &  1 (9.1\%) \\
    Pointed
      &  5 (62.5\%) &  2 (25.0\%) & 0 (0.00\%) &  1 (12.5\%) \\
    Showed items
      &  0 (0.0\%)  &  1 (25.0\%) & 2 (50.0\%) &  0 (0.0\%) \\
    Nodded
      &  1 (25.0\%) &  2 (50.0\%) & 0 (0.0\%)  &  0 (0.0\%) \\
    Touched belongings
      &  0 (0.0\%)  &  1 (33.3\%) & 0 (0.0\%)  &  2 (66.7\%) \\
    \bottomrule
  \end{tabular}
\end{table*}


\section{Nonverbal Cues Detection}
In Section~\ref{sec:datacollect}, we showed that in teleoperated customer-service interactions in a retail store, customers' nonverbal behaviors serve as triggers for starting and sustaining conversations. To translate this into autonomous customer-service dialogue, we must estimate such behaviors from video in real time and output them in a form usable for proactive utterance control (e.g., initiating interaction, asking questions at appropriate moments, or switching topics). Accordingly, in this section, we formulate a recognition model that estimates the following nine behaviors from input video: \textit{Approached}, \textit{Walked away}, \textit{User--user interaction}, \textit{Nodded}, \textit{Showed items}, \textit{Peered into robot}, \textit{Touched belongings}, \textit{Pointed}, and \textit{Waved}.

\label{sec:action_recognition}

\subsection{Spatio-Temporal Nonverbal Cues Recognition}

\begin{figure}[t]
    \centering
    \includegraphics[width=1.00\linewidth]{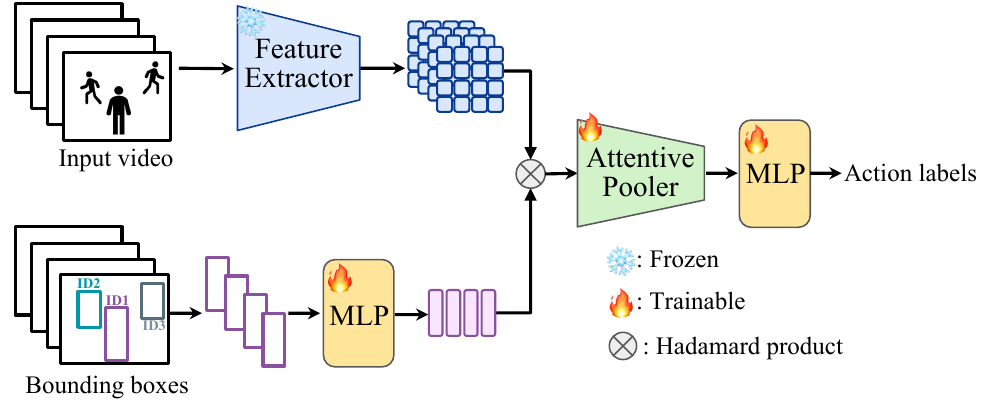}
    \caption{Proposed architecture for user nonverbal behavior recognition}
    \label{fig:proposal_overview}
\end{figure}

An outline of the proposed architecture for user nonverbal behavior recognition is shown in Fig.~\ref{fig:proposal_overview}.

\textbf{Formulation:}
\space The inputs are a video clip $\mathbf{X}=\{I_t\}_{t=1}^{T}$ consisting of $T$ frames and a sequence of bounding boxes $\mathbf{b}_i=\{\mathbf{b}_{i,t}\}_{t=1}^{T}$ for each person $i\in\{1,\dots,P\}$ appearing in the clip.
Here, $\mathbf{b}_{i,t}=(x_{i,t},y_{i,t},w_{i,t},h_{i,t})$ represents the location of person $i$ at frame $t$.
The output is a set of non-mutually-exclusive (multi-label) nonverbal behavior classes for each person $i$, and the supervision is given as a multi-label vector $\mathbf{y}_i=[y_{i,1},\dots,y_{i,C}]^\top\in\{0,1\}^{C}$.

\textbf{Proposed method:}
\space First, we feed the video clip $\mathbf{X}$ into a pretrained feature extractor $\phi(\cdot)$ to obtain visual  scene features.
We use a Vision Transformer as the feature extractor.
\begin{equation}
  \mathbf{f}=\phi(\mathbf{X})\in\mathbb{R}^{N \times D}.
\end{equation}
Here, $N$ is the number of tokens and $D$ is the feature dimension.

Next, we normalize each person's bounding box sequence $\mathbf{b}_i$ by the frame size to obtain $\tilde{\mathbf{b}}_{i,t}$ and form a sequence by concatenating the $T$ frames.
\begin{equation}
  \tilde{\mathbf{b}}_i=
  [\tilde{\mathbf{b}}_{i,1};\dots;\tilde{\mathbf{b}}_{i,T}]
  \in\mathbb{R}^{T \times 4}.
\end{equation}
We feed this sequence into an MLP $\psi(\cdot)$ to extract positional features for person $i$.
\begin{equation}
  \mathbf{p}_{i}=\psi(\tilde{\mathbf{b}}_{i})\in\mathbb{R}^{T \times D}.
\end{equation}
The MLP is shared across frames.

Then, to focus the scene features $\mathbf{f}$ on person $i$, we compute fused features via an element-wise (Hadamard) product between $\mathbf{f}$ and the positional features broadcast along the token dimension.
\begin{equation}
  \mathbf{h}_{i}=\mathbf{f}\odot \mathbf{p}_i\in\mathbb{R}^{N \times D}.
\end{equation}
This operation gates the generic scene features with the positional features $\mathbf{p}_{i}$, transforming them into representations relevant to person $i$.

Next, we aggregate the fused features with an attentive pooler.
The attentive pooler uses a learnable query $\mathbf{q}\in\mathbb{R}^{D}$ and applies multi-head cross-attention to the fused features $\mathbf{h}_i$ to obtain a person-specific representation.
\begin{equation}
  \mathbf{z}_i=\mathrm{CrossAttn}(\mathbf{q},\mathbf{h}_i) \in \mathbb{R}^{D}.
\end{equation}

Finally, we feed $\mathbf{z}_i$ into an MLP classifier $g(\cdot)$ to compute the logits $\mathbf{s}_i$ for each class.
\begin{equation}
  \mathbf{s}_i=g(\mathbf{z}_i)\in\mathbb{R}^{C},\quad
  \hat{y}_{i,c}=\sigma(s_{i,c})
\end{equation}

We train the model using binary cross-entropy (BCE) loss.
Because our dataset is highly imbalanced, we introduce class weights based on the inverse of the number of positive examples $n_c$ for each class $c$.
\begin{equation}
  \mathcal{L}
  =\frac{1}{|\mathcal{B}|}\sum_{i\in\mathcal{B}}\sum_{c=1}^{C}
  w_c\mathrm{BCE}(\hat{y}_{i,c}, y_{i,c}), \quad
  w_c = \frac{1}{n_c}.
\end{equation}
This increases the relative contribution of minority classes and mitigates learning bias.

\subsection{Recognition results}
For the feature extractor, we use V-JEPA2~\cite{assran2025vjepa2}, a ViT-based foundation model for general-purpose visual features. To evaluate the proposed method without confounding errors from person detection or tracking, in this section, we use the dataset's ground-truth bounding boxes and tracking IDs. Each person (bounding box) in every frame is annotated with the corresponding nonverbal behavior labels. For training, we segment videos into clips of $T$ frames and perform action recognition only for persons whose bounding boxes are available throughout all $T$ frames. We construct the training set by sliding the clip start time in steps of $w$ frames. The method is implemented in PyTorch and trained for 20 epochs with batch size 16. We evaluate performance with 5-fold cross-validation and report the averaged results.

When running inference with the trained nonverbal-cue recognizer on a live camera stream, we form an input clip using a sliding window of the most recent eight frames. In our current implementation, the recognizer (excluding person detection/tracking) performs inference at approximately 8 FPS. This satisfies the real-time requirement in this paper (>1 FPS) and enables the dialogue module to condition its responses on recently observed user cues.

As a reference, describing users' nonverbal behaviors in a 2-second video clip took $6.63 \pm 1.19$\,s with Gemini 3.5 Flash and $2.54 \pm 0.58$\,s with Gemini 3.5 Flash Lite on average, both exceeding the one-second response budget; we therefore exclude VLM baselines from our comparison.

Table~\ref{tab:vjepa2_metrics} reports precision, recall, F1, and accuracy for each label. Because the labels are highly imbalanced, we treat micro-F1 and per-class precision/recall as the primary indicators. In terms of F1-score, all labels except \textit{User--user interaction} exceed 0.6, with a micro-average F1 of 0.8. Behaviors with larger motion (e.g., approaching and showing items) are easier to capture and achieve higher scores. For rarer behaviors such as pointing and touching belongings, the F1-score is around 0.6, suggesting that the model can still recognize minority labels in this imbalanced dataset (see Table~\ref{tab:nonverbal-stats} for the label distribution).

\begin{table}[t]
\centering
\caption{Accuracy of nonverbal cues detection}
\label{tab:vjepa2_metrics}
\begin{tabular}{l|cccc}
\toprule
Recognition labels & precision & recall & F1 & accuracy \\
\midrule
Approached & 0.99 & 0.83 & 0.90 & 0.89 \\
Waved & 0.57 & 0.62 & 0.59 & 0.98 \\
Walked away & 0.50 & 0.83 & 0.62 & 0.95 \\
User--user interaction & 0.95 & 0.31 & 0.47 & 0.95 \\
Peered into robot & 0.89 & 0.71 & 0.79 & 0.99 \\
Pointed & 0.57 & 0.67 & 0.62 & 1.00 \\
Showed items & 0.95 & 0.82 & 0.88 & 1.00 \\
Nodded & 0.82 & 0.54 & 0.65 & 0.98 \\
Touched belongings & 0.69 & 0.52 & 0.61 & 0.92 \\
\midrule
micro Avg. & 0.89 & 0.73 & 0.80 & 0.97 \\
\bottomrule
\end{tabular}
\end{table}

\section{Utterance Generation}
We integrate a recognition model for nine key nonverbal behaviors into an utterance-generation system. Because we aim to mimic the content of human customer-service dialogue in this study, we focus on utterance generation rather than utterance timing. Natural turn-taking can be handled by methods such as Voice Activity Projection (VAP)~\cite{inoue2025vap}.

\subsection{Nonverbal Cues-Aware Utterance Generation}

\begin{figure}
    \centering
    \includegraphics[width=1.0\linewidth]{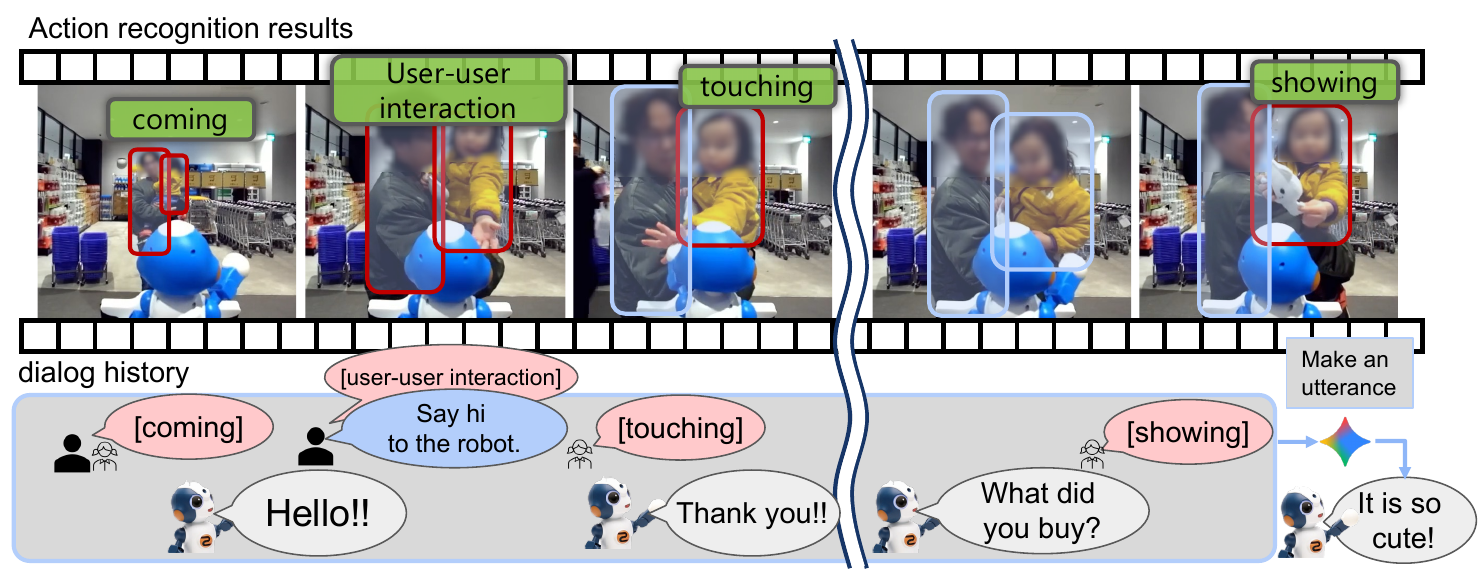}
    \caption{Ideal interaction generated by the proposed system}
    \label{fig:utterance_generation}
\end{figure}

Fig.~\ref{fig:utterance_generation} shows the ideal interaction generated by the proposed system, which conditions utterance generation on nonverbal behaviors. As a base dialogue system, we use a cascaded pipeline with an LLM: the user's speech is transcribed by STT, and the LLM generates the next utterance from the dialogue history~\cite{chen2024dvoiceassit}.

The nonverbal cues detection runs in parallel to the dialogue system. Using camera images and person-detection results, it emits a nonverbal label only when one of the nine target behavior states is observed. A temporal filter suppresses transient false positives. The emitted label is appended to the dialogue history and provided to the LLM, allowing it to generate responses that reflect the user's nonverbal behavior. Only when the \textit{Showed items} behavior is detected, we additionally use a VLM to identify items in the current image and generate an utterance about the presented items. Nonverbal labels are encoded as prompt tokens (e.g., [waved], [pointed]), whose meanings are explained in the initial prompt (e.g., [waved] indicates the user is waving).

Overall, an LLM serves as the dialogue backbone while a lightweight nonverbal recognizer runs concurrently, enabling real-time utterance generation.

\subsection{Offline Experiments}
\label{sec:offline_experiments}
In the collected dataset, we evaluate the proposed utterance generation method for nonverbal-triggered utterance scenes, which cannot be generated by conventional dialogue systems. An interaction video segment and the dialogue history immediately preceding a robot's turn are used to generate the robot's next utterance. Both the video and dialogue history include only information available before the target turn. We examine the extent to which the generated utterances approximate those of the teleoperator.

For evaluation, we map both teleoperator and generated utterances to the intent categories in Table~\ref{tab:trigger-label-stats} and compute their agreement. Intent labels for generated utterances were assigned by human annotators using the same schema as Table II, independent of the teleoperator labels. This intent-level evaluation is adopted because the objective is not to reproduce the teleoperator's exact wording, but to assess whether the system correctly interprets the relevant nonverbal cue and produces an appropriate dialogue intent. We evaluate the 334 nonverbal-triggered utterances in our dataset.

We use ground-truth person detections and dialogue histories because the aim in this section is to assess the accuracy of nonverbal-triggered utterance based on recognition, rather than the end-to-end performance of an online integrated system. The LLM prompt specifies only the robot role, environmental information, and the semantics of nonverbal behavior tokens; we deliberately avoid explicit rules for responding to each label to test performance under a minimal-prompt setting. We use Gemini 2.5 Flash as the LLM.

\begin{figure}
    \centering
    \includegraphics[width=0.65\linewidth]{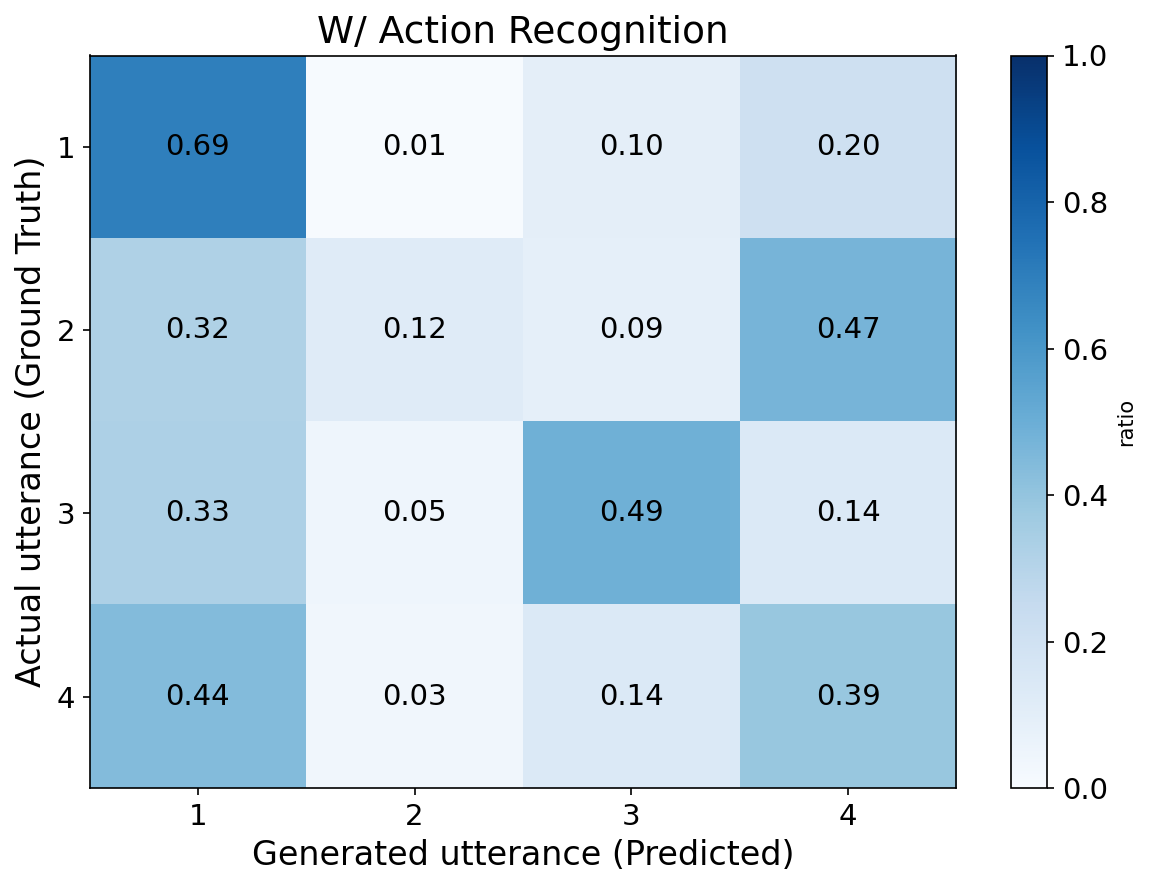}
    \caption{Intent classification results. Label~1 denotes \textit{Social \& Greeting}, label~2 denotes \textit{Task-related info \& directives}, label~3 denotes \textit{User/state comment \& empathy}, and label~4 denotes \textit{Engagement invitation}.}
    \label{fig:intent_matrix}
\end{figure}



For Fig.~\ref{fig:intent_matrix}, we report results on the four main intent classes excluding \textit{Other}. The overall agreement between teleoperator intents and those derived from our generated utterances is 49.1\%, but this aggregates intents with very different prevalence and difficulty. Notably, the dominant \textit{Social \& Greeting} achieves 69\% accuracy, whereas the rare \textit{Task-related info \& directives}—often under-specified under our deliberately minimal prompting—substantially lowers the overall score. A naive majority-class baseline reaches 44.0\%, indicating a consistent improvement from nonverbal conditioning.

These trends are consistent with the correspondence between nonverbal behaviors and intent labels (Table~\ref{tab:nonverbal-cross}). The comparatively high accuracy for \textit{Social \& Greeting} is largely attributable to behaviors such as \textit{Approached}, \textit{Walked away}, \textit{Pointed}, and \textit{Waved}, which are biased toward this intent. Moreover, frequent behaviors such as \textit{Approached} (29.2\%) and \textit{Waved} (11.6\%) (Table~\ref{tab:nonverbal-stats}) provide clear discrete cues in the prompt, facilitating selection of \textit{Social \& Greeting}.

In contrast, the low accuracy for \textit{Task-related info \& directives} can be attributed to the small overall prevalence of this label (Table~\ref{tab:trigger-label-stats}) and its strongest associated behavior being rare: \textit{Touched belongings} (66.7\% in Table~\ref{tab:nonverbal-cross}, but 0.7\% in Table~\ref{tab:nonverbal-stats}). Additionally, utterances in this category often require environment-specific details (e.g., campaign information or cautions) that are difficult to infer under minimal prompting. The teleoperator also occasionally produced such utterances at timings that disregard the surrounding context, further complicating intent inference and increasing misclassification. Consequently, the model tends to over-predict the safer and less content-specific \textit{Engagement invitation}.

Finally, confusion between \textit{Social \& Greeting} and \textit{Engagement invitation} arises because identical nonverbal behaviors can plausibly elicit either intent. Although \textit{Approached}, \textit{Waved}, and \textit{Pointed} are biased toward \textit{Social \& Greeting}, they still account for at least 20\% of \textit{Engagement invitation} instances (Table~\ref{tab:nonverbal-cross}), yielding an inherently ambiguous boundary when conditioning on nonverbal cues alone.

\subsection{Online Prototype: Real-Time Interactive Dialogue System}
\label{sec:online_prototype}

To verify that the proposed recognition-to-generation pipeline can operate online, we implemented a real-time interactive dialogue prototype that connects the trained nonverbal-cue recognizer with LLM-based utterance generation. While offline experiments isolate nonverbal cue recognition using ground-truth tracks, the online prototype uses an automatic detector/tracker to demonstrate feasibility under realistic sensing noise.

\paragraph{System configuration} Our prototype runs on a PC equipped with an Intel Core i7-11800H CPU, 32 GB RAM, and an NVIDIA GeForce RTX 3080 GPU. The system takes camera and microphone streams as input and processes them through a person detection/tracking module, implemented using a custom model based on DEIMv2~\cite{Hyodo2025DEIMv2, huang2025realdeimv2}, which outputs per-frame bounding boxes and track IDs. User speech is transcribed with Google Cloud Speech-to-Text (STT). Based on the dialogue history and recognized nonverbal-cue tokens, the system generates responses using Gemini 2.5 Flash and synthesizes speech using VOICEVOX. A lightweight runtime layer synchronizes these modules and manages inter-process communication.

\paragraph{Qualitative system behavior in supplementary videos} Supplementary videos provide qualitative examples of the proposed integrated system, including proactive greetings elicited by approaching or waving, speech generation in response to a user nod, and cue-conditioned comments emitted when the user shows an item.

\section{Discussion and Limitations}
We presented a framework for customer-service dialogue in a real store, treating users' nonverbal behaviors preceding robot utterances as triggers for interaction initiation and intervention. The framework includes real-world data collection and analysis via teleoperation, selection of nine operationally important behaviors, and real-time recognition of them under multi-person conditions, and proactive utterance generation conditioned on recognition results. Designing recognition targets from the phenomenon of nonverbal-triggered utterances offers practical guidance beyond speech-input-centered dialogue design.

Our approach is not a simple reflex of speaking upon detection; it leverages dialogue history to choose appropriate intent and phrasing during brief contact opportunities. Future work should improve environment-dependent guidance and cautions by incorporating store-specific knowledge, for example, via a two-stage architecture (lightweight intent selection + LLM) that separates intent control from surface realization. Evaluation should also go beyond intent-label agreement by penalizing missing task guidance and incorporating user-experience measures.

This study has several limitations. First, our evaluation is mainly offline and does not fully capture how real-world errors (e.g., person detection/tracking) propagate through the pipeline; online, end-to-end evaluations with user metrics are needed. Second, because the data were collected from teleoperated service at the entrance of a single store, generalization across store types, crowd levels, layouts, and operator strategies is limited, and intent imbalance may affect estimation and learning. Third, we evaluated utterance generation primarily at the intent level, only indirectly addressing phrasing quality, factual correctness, safety, and service quality. Future work should expand data across environments, assess robustness, and combine objective metrics with subjective quality assessments.

\bibliographystyle{ieeetr}
\bibliography{ref.bib}

@article{ravi2024sam2,
  title={SAM 2: Segment Anything in Images and Videos},
  author={Ravi, Nikhila and Gabeur, Valentin and Hu, Yuan-Ting and Hu, Ronghang and Ryali, Chaitanya and Ma, Tengyu and Khedr, Haitham and R{\"a}dle, Roman and Rolland, Chloe and Gustafson, Laura and Mintun, Eric and Pan, Junting and Alwala, Kalyan Vasudev and Carion, Nicolas and Wu, Chao-Yuan and Girshick, Ross and Doll{\'a}r, Piotr and Feichtenhofer, Christoph},
  journal={arXiv preprint arXiv:2408.00714},
  year={2024}
}

@inproceedings{yano2024unified,
  title={Unified understanding of environment, task, and human for human-robot interaction in real-world environments},
  author={Yano, Yuga and Mizutani, Akinobu and Fukuda, Yukiya and Kanaoka, Daiju and Ono, Tomohiro and Tamukoh, Hakaru},
  booktitle={IEEE International Conference on Robot and Human Interactive Communication},
  pages={224--230},
  year={2024},
}

@article{hurst2024gpt4o,
  title={GPT-4o system card},
  author={Hurst, Aaron and Lerer, Adam and Goucher, Adam P and Perelman, Adam and Ramesh, Aditya and Clark, Aidan and Ostrow, AJ and Welihinda, Akila and Hayes, Alan and Radford, Alec and others},
  journal={arXiv preprint arXiv:2410.21276},
  year={2024}
}

@inproceedings{bunt2020iso,
  title     = {The ISO Standard for Dialogue Act Annotation, Second Edition},
  author    = {Bunt, Harry and Petukhova, Volha and Gilmartin, Emer and Pelachaud, Catherine and Fang, Alex and Keizer, Simon and Pr{\'e}vot, Laurent},
  booktitle = {International Conference on Language Resources and Evaluation},
  pages     = {549--558},
  year      = {2020},
}

@article{chen2017survey,
  title     = {A Survey on Dialogue Systems: Recent Advances and New Frontiers},
  author    = {Chen, Hongshen and Liu, Xiaorui and Yin, Dawei and Tang, Jiliang},
  journal   = {ACM SIGKDD Explorations Newsletter},
  volume    = {19},
  number    = {2},
  pages     = {25--35},
  year      = {2017},
  publisher = {Association for Computing Machinery}
}

@inproceedings{nothdurft2015proactive,
  title     = {Finding Appropriate Interaction Strategies for Proactive Dialogue Systems---An Open Quest},
  author    = {Nothdurft, Florian and Ultes, Stefan and Minker, Wolfgang},
  booktitle = {European and the 5th Nordic Symposium on Multimodal Communication},
  pages     = {73--80},
  year      = {2015},
}

@article{tisserand2024unraveling,
  title   = {Unraveling the Thread: Understanding and Addressing Sequential Failures in Human--Robot Interaction},
  author  = {Tisserand, Lucien and Stephenson, Brooke and Baldauf-Quilliatre, Heike and Lefort, Mathieu and Armetta, Fr{\'e}d{\'e}ric},
  journal = {Frontiers in Robotics and AI},
  volume  = {11},
  pages   = {1359782},
  year    = {2024},
  doi     = {10.3389/frobt.2024.1359782}
}

@article{huang2025realdeimv2,
  title={Real-Time Object Detection Meets DINOv3},
  author={Huang, Shihua and Hou, Yongjie and Liu, Longfei and Yu, Xuanlong and Shen, Xi},
  journal={arXiv preprint arXiv:2509.20787},
  year={2025}
}

@article{comanici2025gemini,
  title={Gemini 2.5: Pushing the frontier with advanced reasoning, multimodality, long context, and next generation agentic capabilities},
  author={Comanici, Gheorghe and Bieber, Eric and Schaekermann, Mike and Pasupat, Ice and Sachdeva, Noveen and Dhillon, Inderjit and Blistein, Marcel and Ram, Ori and Zhang, Dan and Rosen, Evan and others},
  journal={arXiv preprint arXiv:2507.06261},
  year={2025}
}

@inproceedings{feichtenhofer2019slowfast,
  title={Slowfast networks for video recognition},
  author={Feichtenhofer, Christoph and Fan, Haoqi and Malik, Jitendra and He, Kaiming},
  booktitle={IEEE/CVF international conference on computer vision},
  pages={6202--6211},
  year={2019}
}

@article{assran2025vjepa2,
  title={V-JEPA 2: Self-supervised video models enable understanding, prediction and planning},
  author={Assran, Mido and Bardes, Adrien and Fan, David and Garrido, Quentin and Howes, Russell and Muckley, Matthew and Rizvi, Ammar and Roberts, Claire and Sinha, Koustuv and Zholus, Artem and others},
  journal={arXiv preprint arXiv:2506.09985},
  year={2025}
}

@ARTICLE{song2025robotclerk,
  author={Song, Sichao and Iwamoto, Takuya and Okafuji, Yuki and Baba, Jun and Nakanishi, Junya and Yoshikawa, Yuichiro and Ishiguro, Hiroshi},
  journal={IEEE Robotics and Automation Letters}, 
  title={From Attraction to Engagement: A Robot-Clerk Collaboration Strategy for Retail Success}, 
  year={2025},
  volume={10},
  number={7},
  pages={6672-6679},
  doi={10.1109/LRA.2025.3568611}}

@INPROCEEDINGS{amy2025hri_motivation,
  author={Koike, Amy and Okafuji, Yuki and Hoshimure, Kenya and Baba, Jun},
  booktitle={ACM/IEEE International Conference on Human-Robot Interaction}, 
  title={What Drives You to Interact?: The Role of User Motivation for a Robot in the Wild}, 
  year={2025},
  pages={183-192},
  doi={10.1109/HRI61500.2025.10974089}
}

@InProceedings{pmlr25ex-vad,
  title = 	 {Ex-{VAD}: Explainable Fine-grained Video Anomaly Detection Based on Visual-Language Models},
  author =       {Huang, Chao and Shi, Yushu and Wen, Jie and Wang, Wei and Xu, Yong and Cao, Xiaochun},
  booktitle = 	 {Proceedings of the 42nd International Conference on Machine Learning},
  pages = 	 {25750--25761},
  year = 	 {2025},
  volume = 	 {267},
  series = 	 {Proceedings of Machine Learning Research},
  month = 	 {13--19 Jul},
  publisher =    {PMLR},
}

@article{xu2025streamingvlm,
  title={StreamingVLM: Real-time understanding for infinite video streams},
  author={Xu, Ruyi and Xiao, Guangxuan and Chen, Yukang and He, Liuning and Peng, Kelly and Lu, Yao and Han, Song},
  journal={arXiv preprint arXiv:2510.09608},
  year={2025}
}

@article{wang2021actionclip,
  title={ActionCLIP: A new paradigm for video action recognition},
  author={Wang, Mengmeng and Xing, Jiazheng and Liu, Yong},
  journal={arXiv preprint arXiv:2109.08472},
  year={2021}
}

@article{urakami2023nonverbal,
    author = {Urakami, Jacqueline and Seaborn, Katie},
    title = {Nonverbal Cues in Human–Robot Interaction: A Communication Studies Perspective},
    year = {2023},
    publisher = {Association for Computing Machinery},
    address = {New York, NY, USA},
    volume = {12},
    number = {2},
    url = {https://doi.org/10.1145/3570169},
    doi = {10.1145/3570169},
    journal = {Journal of Humam-Robot Interaction},
    articleno = {22},
    numpages = {21}
}

@INPROCEEDINGS{chamoto2023task,
  author={Chamoto, Yuki and Okafuji, Yuki and Matsumura, Kohei and Baba, Jun and Nakanishi, Junya},
  booktitle={IEEE International Conference on Robot and Human Interactive Communication}, 
  title={Investigating the Influence of Task-dependent and Task-independent Robot Behavior on the Impression of Robots and the User Experience}, 
  year={2023},
  volume={},
  number={},
  pages={640-646},
  doi={10.1109/RO-MAN57019.2023.10309453}
}

@article{neef2023appropriate,
  author  = {Nicolas E. Neef and Sarah Zabel and Mathis Lauckner and Siegmar Otto },
  title   = {What is Appropriate? On the Assessment of Human-Robot Proxemics for Casual Encounters in Closed Environments},
  journal = {International Journal of Social Robotics},
  volume  = {15},
  pages   = {953--967},
  year    = {2023},
  doi     = {10.1007/s12369-023-01004-1},
  url     = {https://doi.org/10.1007/s12369-023-01004-1}
}

@inproceedings{skantze2025turntaking,
    author = {Skantze, Gabriel and Irfan, Bahar},
    title = {Applying General Turn-taking Models to Conversational Human-Robot Interaction},
    year = {2025},
    booktitle = {ACM/IEEE International Conference on Human-Robot Interaction},
    pages = {859–868},
    numpages = {10},
    location = {Melbourne, Australia},
}

@inproceedings{constantin2023interactive,
  title        = {Interactive Multimodal Robot Dialog Using Pointing Gesture Recognition},
  author       = {Constantin, Stefan and Eyiokur, Fevziye Irem and Yaman, Dogucan and B{\"a}rmann, Leonard and Waibel, Alex},
  booktitle    = {European Computer Vision Association Workshops},
  pages        = {640--657},
  year         = {2023},
  doi          = {10.1007/978-3-031-25075-0_43},
  url          = {https://link.springer.com/chapter/10.1007/978-3-031-25075-0_43}
}

@inproceedings{axelsson2023doyoufollow,
  title        = {Do You Follow?: A Fully Automated System for Adaptive Robot Presenters},
  author       = {Axelsson, Agnes and Skantze, Gabriel},
  booktitle    = {ACM/IEEE International Conference on Human-Robot Interaction},
  pages        = {102--111},
  year         = {2023},
  doi          = {10.1145/3568162.3576958},
  url          = {https://doi.org/10.1145/3568162.3576958}
}

@article{allgeuer2024chatty,
  title={When Robots Get Chatty: Grounding Multimodal Human-Robot Conversation and Collaboration},
  author={Allgeuer, Philipp and Ali, Hassan and Wermter, Stefan},
  journal={arXiv preprint arXiv:2407.00518},
  year={2024}
}

@INPROCEEDINGS{inoue2025vap,
  author={Inoue, Koji and Okafuji, Yuki and Baba, Jun and Ohira, Yoshiki and Hyodo, Katsuya and Kawahara, Tatsuya},
  booktitle={IEEE/RSJ International Conference on Intelligent Robots and Systems}, 
  title={A Noise-Robust Turn-Taking System for Real-World Dialogue Robots: A Field Experiment}, 
  year={2025},
  volume={},
  number={},
  pages={874-879},
  doi={10.1109/IROS60139.2025.11246533}}

@inproceedings{zhao2022tuber,
  title={Tuber: Tubelet transformer for video action detection},
  author={Zhao, Jiaojiao and Zhang, Yanyi and Li, Xinyu and Chen, Hao and Shuai, Bing and Xu, Mingze and Liu, Chunhui and Kundu, Kaustav and Xiong, Yuanjun and Modolo, Davide and others},
  booktitle={IEEE/CVF Conference on Computer Vision and Pattern Recognition},
  pages={13598--13607},
  year={2022}
}

@article{okafuji2024eat,
    author = {Yuki Okafuji and Takumi Ishikawa and Kohei Matsumura and Jun Baba and Junya Nakanishi},
    title = {Pseudo-eating behavior of service robot to improve the trustworthiness of product recommendations},
    journal = {Advanced Robotics},
    volume = {38},
    number = {5},
    pages = {343--356},
    year = {2024},
    publisher = {Taylor \& Francis},
    doi = {10.1080/01691864.2024.2321191},
    URL = {https://doi.org/10.1080/01691864.2024.2321191},
    eprint = {https://doi.org/10.1080/01691864.2024.2321191}
}

@inproceedings{girdhar2019video,
  title={Video action transformer network},
  author={Girdhar, Rohit and Carreira, Joao and Doersch, Carl and Zisserman, Andrew},
  booktitle={IEEE/CVF conference on computer vision and pattern recognition},
  pages={244--253},
  year={2019}
}

@inproceedings{yuan2025videorefer,
  title={VideoRefer suite: Advancing spatial-temporal object understanding with video LLM},
  author={Yuan, Yuqian and Zhang, Hang and Li, Wentong and Cheng, Zesen and Zhang, Boqiang and Li, Long and Li, Xin and Zhao, Deli and Zhang, Wenqiao and Zhuang, Yueting and others},
  booktitle={Computer Vision and Pattern Recognition Conference},
  pages={18970--18980},
  year={2025}
}

@article{chen2024dvoiceassit,
  title={VoiceBench: Benchmarking LLM-Based Voice Assistants},
  author={Yiming Chen and Xianghu Yue and Chen Zhang and Xiaoxue Gao and Robby T. Tan and Haizhou Li},
  journal={arXiv preprint arXiv:2410.17196},
  year={2024}
}

@article{mitchel2011uncanny,
    author = {Wade J Mitchell and Kevin A Szerszen, Sr and Amy Shirong Lu and Paul W Schermerhorn and Matthias Scheutz and Karl F MacDorman},
    title ={A Mismatch in the Human Realism of Face and Voice Produces an Uncanny Valley},
    journal = {i-Perception},
    volume = {2},
    number = {1},
    pages = {10-12},
    year = {2011},
    doi = {10.1068/i0415},
    URL = {https://doi.org/10.1068/i0415},
    eprint = {https://doi.org/10.1068/i0415},
}

@article{nyatsanga2023gesture,
  title={A Comprehensive Review of Data-Driven Co-Speech Gesture Generation},
  author={Simbarashe Nyatsanga and Taras Kucherenko and Chaitanya Ahuja and Gustav Eje Henter and Michael Neff},
  journal={arXiv preprint arXiv:2301.05339},
  year={2023}
}

@inproceedings{satake2009approach,
    author = {Satake, Satoru and Kanda, Takayuki and Glas, Dylan F. and Imai, Michita and Ishiguro, Hiroshi and Hagita, Norihiro},
    booktitle={ACM/IEEE International Conference on Human-Robot Interaction}, 
    title = {How to approach humans? strategies for social robots to initiate interaction},
    year = {2009},
    isbn = {9781605584041},
    url = {https://doi.org/10.1145/1514095.1514117},
    doi = {10.1145/1514095.1514117},
}

@INPROCEEDINGS{Shiwa2008quick,
  author={Shiwa, Toshiyuki and Kanda, Takayuki and Imai, Michita and Ishiguro, Hiroshi and Hagita, Norihiro},
  booktitle={ACM/IEEE International Conference on Human-Robot Interaction}, 
  title={How quickly should communication robots respond?}, 
  year={2008},
  volume={},
  number={},
  pages={153-160},
  doi={10.1145/1349822.1349843}
}

@misc{Hyodo2025DEIMv2,
  author = {Katsuya Hyodo},
  title  = {472\_DEIMv2-Wholebody34: Lightweight human detection models generated on high-quality human data sets},
  year   = {2025},
  note   = {Available: \url{https://github.com/PINTO0309/PINTO_model_zoo/tree/main/472_DEIMv2-Wholebody34}}
}

\end{document}